\documentclass[%
 aip,
 amsmath,amssymb,
 preprint,%
]{revtex4-1}

\usepackage{graphicx}
\usepackage{dcolumn}
\usepackage{bm}

\usepackage[utf8]{inputenc}
\usepackage[T1]{fontenc}
\usepackage{mathptmx}
\usepackage{algorithm}
\usepackage{algpseudocode}
\usepackage{multirow}  
\usepackage{booktabs}  
\usepackage{siunitx}    

\makeatletter
\def\@email#1#2{%
 \endgroup
 \patchcmd{\titleblock@produce}
  {\frontmatter@RRAPformat}
  {\frontmatter@RRAPformat{\produce@RRAP{*#1\href{mailto:#2}{#2}}}\frontmatter@RRAPformat}
  {}{}
}%
\makeatother
\begin{document}

\preprint{AIP/123-QED}

\title[CbLDM: A Diffusion Model for recovering nanostructure from atomic pair distribution functions]{CbLDM: A Diffusion Model for recovering nanostructure \\ from atomic pair distribution functions}
\author{Jiarui Cao}
 \thanks{These authors contributed equally.}
 \affiliation{School of Statistics and Data Science, Nankai University, Tianjin 300071, China}
\author{Zhiyang Zhang}%
 \thanks{These authors contributed equally.}
\affiliation{NITFID, School of Statistics and Data Science, Nankai University, Tianjin 300071, China}%

\author{Heming Wang}
 \thanks{These authors contributed equally.}
\affiliation{School of Statistics and Data Science, Nankai University, Tianjin 300071, China}%

\author{Jun Xu}
\affiliation{School of Statistics and Data Science, Nankai University, Tianjin 300071, China}%
\author{Ling Lan}
\affiliation{Department of Applied Physics and Applied Mathematics, Columbia University, New York, NY, 10027, USA}%
\author{Simon J. L. Billinge}
\affiliation{Department of Materials, University of California, Santa Barbara, Santa Barbara, CA 93101}%
\author{Ran Gu}
 \email{Author to whom correspondence should be addressed: rgu@nankai.edu.cn}
\affiliation{NITFID, School of Statistics and Data Science, Nankai University, Tianjin 300071, China}%

\date{\today}

\begin{abstract}
The nanostructure inverse problem is an attractive problem that helps researchers to understand the relationship between the properties and the structure of nanomaterials. This study focuses on the problem of recovering the model system of monometallic nanoparticles (MMNPs) from their pair distribution function (PDF) and regards it as a highly ill-posed conditional generation task. This study proposes a Condition-based Latent Diffusion Model (CbLDM) as a feasible solution to this problem. This model demonstrates an acceleration approach within the framework of a latent diffusion model by using conditional priors to estimate the conditional posterior distribution, which is an approximate distribution of $p(z|x)$. In addition, this study uses Laplacian matrix instead of distance matrix to recover the nanostructure, which helps to improve stability. Our study demonstrates that a latent diffusion model with a conditional prior can generate nanostructures that are consistent with PDF observations and physically meaningful, thereby laying the groundwork for subsequent more complex inverse problems.
\end{abstract}

\maketitle

\section{Introduction}
In recent years, nanomaterials have become an important research direction in materials science~\cite{gupta2019carbon}. Due to the size of nanomaterials being close to the electronic coherence length~\cite{billinge2019nanometre}, they exhibit surface effects~\cite{wu2013nanostructure}, small size effects and macroscopic quantum tunneling effects \cite{92Macroscopic}. These features give nanomaterials extensive application prospects in fields such as electronics, optics, and magnetism~\cite{montero2002nanostructures,cho2010effects,yang2008photocurrent,an2009synthesis}. In materials science, nanomaterials display a variety of properties, with one key reason being their structural diversity~\cite{xia2003one,hu2013ultrathin}. Consequently, understanding the relationship between structure and properties motivates the study of the nanostructure inverse problem, which aims to infer structural characteristics from observed properties or behaviors~\cite{billinge2007problem,billinge2010nanostructure}.

Among various experimental techniques, the atomic pair distribution function (PDF) plays an important role in addressing this problem. The advantage of the PDF method over traditional X-ray diffraction (XRD) lies in its ability to integrate all scattering information into the Fourier transform, providing information about atomic distances in nanostructures. This makes PDF particularly suitable for systems with local order, disorder, or finite size effects, such as amorphous materials and nanoparticles~\cite{fernandez2008x}.


However, recovering a three dimensional atomic structure from one dimensional PDF data, constitutes a highly ill-posed inverse problem. A single PDF contains only statistical information about pairwise atomic distances, and distinct atomic configurations may produce very similar PDFs. Experimental noise, finite real-space resolution, and limited measurement ranges further exacerbate the non-uniqueness and instability of the reconstruction. As a result, PDF-based structure recovery is more naturally framed as a probabilistic and generative task, rather than a deterministic one-to-one mapping.

Traditional methods for the nanostructure inverse problem from PDF data include ab initio methods such as the LIGA algorithm~\cite{juhas2006ab,juhas2008liga} and the TRIBOND algorithm~\cite{duxbury2016unassigned}, and the Reverse Monte Carlo method~\cite{mcgreevy2001reverse,soper2000radial}. However, these methods have certain limitations. For example, the TRIBOND algorithm is only suitable for small structures with a small number of atoms, and the time complexity is too high for complex structures, while the LIGA algorithm is almost only suitable for highly symmetric structures~\cite{duxbury2016unassigned}. The reverse Monte Carlo method is efficient only in highly disordered systems, and is inefficient in solving problems with nearly unique solutions~\cite{billinge2008nanoscale}. These limitations become increasingly pronounced for nanostructures containing more than 100 atoms. To address these limitations, it is of interest to explore alternative modeling frameworks that may offer improved flexibility and conceptual generality.

With the increasing application of artificial intelligence, deep learning methods have been introduced to explore alternative solution to the nanostructure inverse problem. The DeepStruc model~\cite{kjaer2023deepstruc}, based on the Conditional Variational Autoencoder (CVAE)~\cite{kingma2013auto}, can reconstruct nanostructures from PDFs. Additionally, a classification model called MlstructureMining has been proposed, which obtains potential nanostructure lists by comparing experimental PDFs with simulated PDFs from a database~\cite{kjaer2024mlstructuremining}. This model is based on the XGBoost ensemble learning algorithm. Although it has shown strong performance in classification tasks~\cite{wu2021comparison}, it encounters challenges in generation tasks where deep learning models provide a more natural formulation. Recently, PXRDnet~\cite{guo2025ab} shows that diffusion model is effective in solving the nanostructure inverse problem. It uses the diffusion model to recover nanostructure with up to 20 atoms in the cell, but the method is mainly for periodic crystal structure. In contrast, PXRDnet uses Powder X-Ray Diffraction(PXRD) to recover nanostructure, and we use PDF data to recover nanostructures. Moreover, our study explores the applicability of diffusion model to amorphous material and nanostructure with larger numbers of atoms.


In this work, we focus on recovering the model system of monometallic nanoparticles (MMNPs) from PDF data and formulate the problem as a highly ill-posed conditional generation task. We propose a Condition-based Latent Diffusion Model (CbLDM) as a solution that combines ideas from conditional variational autoencoder and latent diffusion model~\cite{rombach2022high}. Although DeepStruc has been applied to related structure restoration tasks, it suffers from the problem of generating ambiguity. To address these drawbacks, our CbLDM introduces some innovative designs: by embedding a conditional prior within the latent diffusion framework, the model can generate nanostructures that are consistent with PDF observations and physically meaningful. This study primarily aims to demonstrate the feasibility of using CbLDM for the inverse problem of MMNPs. Additional designs, such as accelerated sampling strategies and the use of Laplacian matrix as the representation of structure, are further introduced to enhance the stability and practicality of the proposed conceptual framework.

\section{Condition-based Latent Diffusion Model}
The nanostructure inverse problem we focus on is a typical conditional generation task, where the goal is to recover the nanostructure given the PDF. In this section, we introduce a conditional latent diffusion framework, Condition-based Latent Diffusion Model (CbLDM), as a solution for the nanostructure inverse problem based on PDF data. In our solution, we first transform the nanostructure inverse problem, which is mathematically an unassigned distance geometry problem (uDGP)~\cite{billinge2016assigned, duxbury2016unassigned}, into an assigned distance geometry problem (aDGP) through CbLDM. That is, directly recovering nanostructures from PDF is transformed into generating a Laplacian distance matrix from PDF. Subsequently, the Laplacian matrix is used to solve the aDGP problem using optimization methods. The CbLDM explores how conditional information can be integrated into latent diffusion frameworks to enable structure determination under severely ill-posed conditions. (The concepts mentioned in this paper, including PDF, uDGP, aDGP, the Laplacian distance matrix, CVAE and DDM will be introduced in detail in the appendix.)
\subsection{CbLDM Framework}\label{framework}
\begin{figure}[!htbp]
\centering 
\includegraphics[width=17cm]{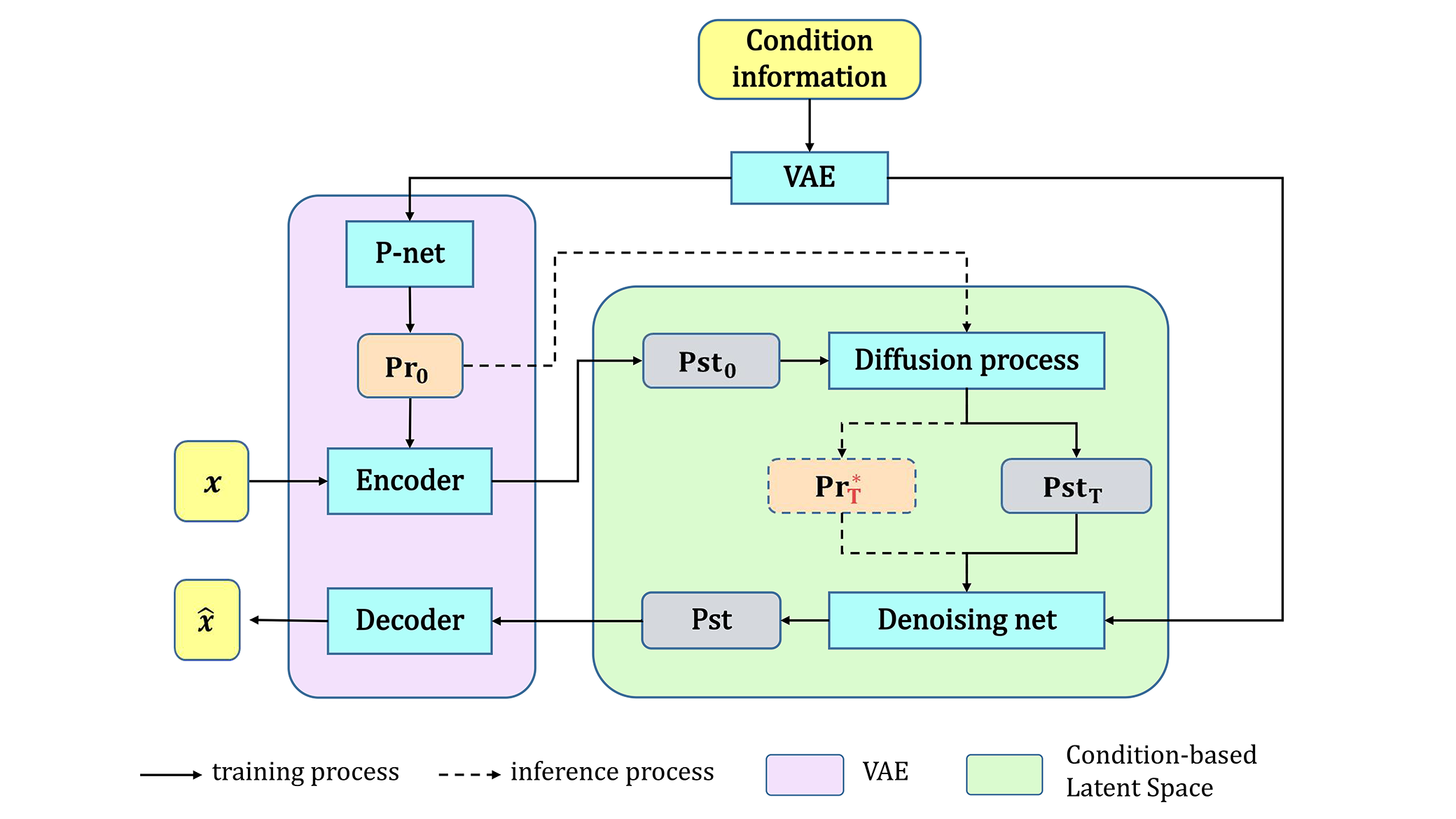}
\caption{CbLDM Model Architecture} 
\label{fig:CbLDM} 
\end{figure}

The overall architecture of CbLDM is illustrated in Fig.~\ref{fig:CbLDM}. CbLDM consists of three functional components: a condition embedding module, a VAE constructing latent space, and a diffusion model operating within this latent space.

The first component is a condition embedding module, whose purpose is to transform the input conditions into a compressed representation. This module is not restricted to a specific architecture and may be implemented using any model capable of extracting distinguishing features from the condition information (in this study, a VAE is employed). For continuous conditions, the embedding is designed to vary smoothly with the input in order to preserve consistency with the original condition space. When the condition information is already represented as low-dimensional vectors, this module may reduce to an identity mapping. This flexible design reflects the exploratory nature of CbLDM.

The second component is a VAE that constructs a latent space. Unlike standard VAEs that adopt an unconditional standard normal prior, the VAE in CbLDM employs a conditional prior distribution. This design allows the latent representations to exhibit partial consistency with both the structural data and the condition information. Importantly, the conditional information is incorporated in the encoder; the decoder itself remains unconditional and reconstructs data solely from latent variables. This choice distinguishes CbLDM from conventional CVAE and allows the latent space structure to be constructed implicitly by the conditions.

The third component is a diffusion model trained in the latent space. Its role is to model the distribution of latent representations based on the embedded conditions and to generate samples from noise. The diffusion model serves as a generative mechanism that explores how conditional information can guide sampling in a low-dimensional latent space.

In our implementation, 1D-convolutional networks are employed for condition embedding and latent space construction, with architectural asymmetry between the encoder and decoder to accommodate the differing roles of compression and reconstruction. 

\subsection{Model Settings}
For the three components of the model introduced in Section~\ref{framework}, we present one representative training strategy adopted to validate the proposed framework in this subsection. As illustrated in Fig.~\ref{fig:CbLDM}, the overall training procedure is organized into three stages corresponding to the three components of the model, with the procedures summarized in Alg.1, Alg.2, and Alg.3.

First, we train the model used to encode and compactly represent the conditional information. As the compressed conditional information will be utilised in the subsequent two stages, this step aims to facilitate the use of the resulting features as a unified conditional reference for both the Variational Autoencoder (VAE) and the diffusion model. In our implementation, the PDF data are represented as tensors of shape $(6,500)$ and encoded into latent vectors of shape $(2,125)$. These dimensional choices are made primarily for numerical stability and computational convenience. As shown in Alg.1, $c$ denotes the conditional input, which is PDF data in our work, and $z_c$ denotes its corresponding latent representation. The loss function of the VAE consists of a reconstruction term ($L_{\mathrm{rec}}$) and a KL-divergence between the assumed conditional prior and the data posterior, both modeled as Gaussian distributions~\cite{higgins2017beta,burgess2018understanding}. Here, $D_c$ and $E_c$ indicate Decoder and Encoder of VAE, $q(\cdot)$ represents prior distribution, and $p_{\theta}$ represents learnable posterior distribution.

\begin{figure}
    \centering
    \includegraphics[width=1\linewidth]{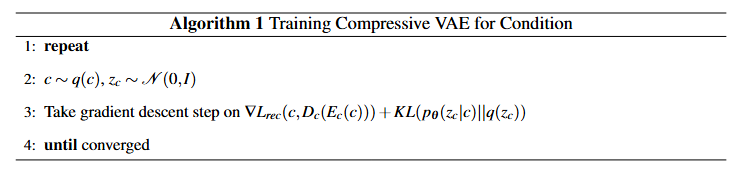}
    \label{al:trainingcofvae}
\end{figure}
After the compressive VAE for the conditional information has converged, we proceed to train the VAE responsible for modeling the input data. To facilitate the modeling of long-range correlations in the Laplacian matrix, the original $(256,256)$ matrix is partitioned into four blocks and rearranged as a $(4,128,128)$ tensor. This representation is compatible with the multi-attention layer employed in the network. The Encoder subsequently maps the input into a low-dimensional latent representation of shape $(1,16,16)$. As shown in Alg.2, $x$ denotes the input data (the Laplacian matrix), $c_0$ denotes the compressed conditional feature obtained from the Encoder of compressed VAE for condition, and $z_x$ denotes the resulting latent representation. The loss function follows the same formulation as that used for the conditional compressive VAE, consisting of a reconstruction term and a KL-divergence regularization. For this implementation, the Adan optimizer~\cite{xie2024adan} is employed for training two VAEs. 

\begin{figure}
    \centering
    \includegraphics[width=1\linewidth]{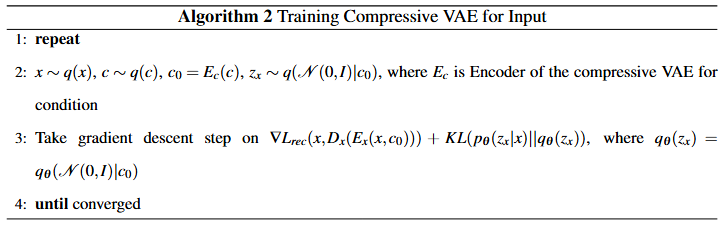}
    \label{al:training input} 
\end{figure}
After both VAEs have converged, a denoising diffusion model (DDM) is trained in the latent space defined by the input compressive VAE. Following the standard formulation of denoising diffusion probabilistic models~\cite{ho2020denoising}, Gaussian noise is added to latent embedding drawn from the conditional posterior, and the diffusion model is trained to predict the added noise. A U-Net architecture is adopted as the backbone of the latent space diffusion model to parameterize the denoising process over the $(1,16,16)$ latent representations. As shown in Alg.3, $z_0$ denotes a latent embedding drawn from the conditional posterior learned by the input compressive VAE. All remaining diffusion hyperparameters follow standard settings commonly used in DDM implementations. The diffusion model is trained using an $L_1$ loss between the added noise and the predicted noise. For this implementation, the AdamW optimizer~\cite{loshchilov2017decoupled} is employed to train the diffusion model.

\begin{figure}
    \centering
    \includegraphics[width=1\linewidth]{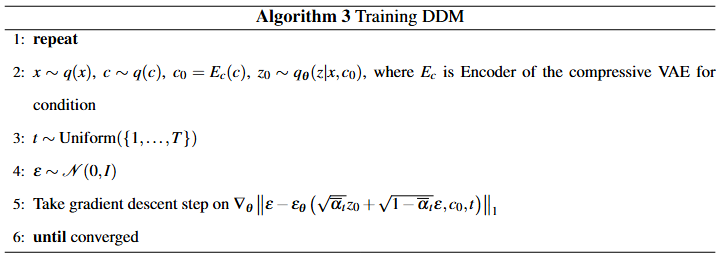}
    \label{al:training ddm} 
\end{figure}

\subsection{Sampling Step}\label{sampling step}
To generate new data from simple distribution, Diffusion Model approximate reverse diffusion process through simulating the process of solving a stochastic differential equation~\cite{song2021maximum}. In standard implementations, sampling typically requires simulating the full extent of reverse diffusion process which can be computationally demanding. Motivated by this observation, we explore a sampling strategy within the CbLDM framework. As shown in Fig.~\ref{fig:CbLDM} and Alg.3, the sampling process is divided into three steps.

\begin{figure}
    \centering
    \includegraphics[width=1\linewidth]{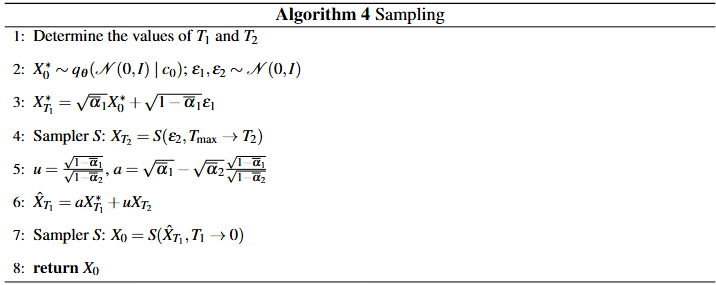}
    \label{al:sampling}
\end{figure}
The first step determines two diffusion time indices $T_1$ and $T_2$ according to different proportion of the noise added to training data in different time. Here, $T_1$ denotes the time step at which a conditional prior sample is noised while $T_2$ denotes the time step reached by denoising a Gaussian noise sample. In the second step, two latent samples are obtained: a sample $X_{T_2}$ obtained by denoising Gaussian noise to time $T_2$, and a conditional prior sample $X^*_{T_1}$ obtained by diffusing a conditional draw to time $T_1$. These two samples are then combined using Eq.~\eqref{equ:estimate XT1} to construct $\hat{X}_{T_1}$ which estimates the sample $X{T_1}$ denoised from Gaussian noise. The last step is to sample final sample from $\hat{X}_{T_1}$ with any sampler $S$ of Diffusion Model. This formulation illustrates how conditional prior information can be integrated into the sampling process, allowing generation to be initialized closer to condition consistent regions of the latent space.
\begin{equation}
    \hat{X}_{T_1} = (\sqrt{\overline{\alpha}_1} - \sqrt{\overline{\alpha}_2} \frac{\sqrt{1-\overline{\alpha}_1}}{\sqrt{1-\overline{\alpha}_2}})X^*_{T_1} + \frac{\sqrt{1-\overline{\alpha}_1}}{\sqrt{1-\overline{\alpha}_2}}X_{T_2}
    \label{equ:estimate XT1}
\end{equation}
Here, the notations used in Alg.3 and Eq.~\eqref{equ:estimate XT1} are defined as follows. $X^*_0$ denotes a sample drawn from the conditional prior~$q_{\theta}(\cdot \mid c)$. $X_{T_1}^*$ denotes that $X^*_0$ diffuses from time 0 to $T_1$. $X_{T_2}$ denotes a latent sample obtained by denoising Gaussian noise up to time step $T_2$. $\epsilon_1$ and $\epsilon_2$ are independent Gaussian noise, and $\overline{\alpha}_1$ and $\overline{\alpha}_2$ are the proportions of data at diffusion process which is similar to DDM~\cite{ho2020denoising}. Under the assumption that the VAE is sufficiently well trained, the learned conditional posterior $p_{\theta}(\cdot \mid c)$ is expected to provide a reasonable approximation to the underlying conditional distribution. Under the assumption that the VAE is sufficiently well trained, we suppose that the conditional approximate posterior~$p_{\theta}(\cdot|c)$ is approximately equal to the corresponding truth posterior~$p(\cdot|c)$. Furthermore, assuming that both the conditional prior~$q_{\theta}(\cdot \mid c)$ and the learned conditional posterior~$p_{\theta}(\cdot \mid c)$ follow Gaussian distributions, their discrepancy can be characterized by a residual distribution with finite mean and variance. Based on the above considerations, the difference between~$q_{\theta}(\cdot|c)$ and~$p(\cdot|c)$ can be abstracted as a distribution~$U$ with finite mean and variance.
\begin{equation}
    U^* = (\sqrt{\overline{\alpha}_1} - \sqrt{\overline{\alpha}_2}\frac{\sqrt{1-\overline{\alpha}_1}}{\sqrt{1-\overline{\alpha}_2}})U.
    \label{equ:explan2}
\end{equation}
Using Eq.~\eqref{equ:estimate XT1}, an estimate of $X_{T_1}$ can be obtained, and the corresponding discrepancy $U^*$ between the estimated and reference samples is characterized by Eq.~\eqref{equ:explan2}. We assume that a sufficiently trained Diffusion Model has the ability to approximate the correct noise distribution, which can cover the $U^*$ with enough small interval between $T_2$ to $T_1$. In practice, the choice of the interval between $T_2$ and $T_1$ is empirically determined based on validation experiments on simulated datasets. In the limiting case where the VAE is not perfect enough, setting $T_1 = T_2$ reduces the procedure to standard latent diffusion sampling. Conversely, in the idealized scenario of a perfect VAE, generation could be initialized directly from conditional prior.

\subsection{Algorithm for recovering structures}

As discussed above, the uDGP is first reformulated into an aDGP through CbLDM. Specifically, the model generates Laplacian matrix conditioned on the input PDF data. The direct recovery of three-dimensional atomic coordinates is not a learning objective of the model. Instead, mapping the generated Laplacian matrix to atomic coordinates is formulated as a subsequent aDGP, which can be addressed using established numerical techniques.

To obtain a concrete three-dimensional realization, the generated Laplacian matrix is first symmetrized and treated as a new Laplacian distance matrix. The eigenvectors associated with the three smallest non-zero eigenvalues are then extracted to construct an initial three-dimensional embedding, serving as an initialization for further refinement. Finally, the MSE of the true and predicted values of the Laplacian matrix is minimized using the minimize function in the scipy library~\cite{virtanen2020scipy}, employing the trust region constrained optimization method~\cite{conn2000trust}.

\section{Numerical Test}
\subsection{Data Set}

All data used in this study are synthetically generated using the DiffPy-CMI library~\cite{juhas2015complex} and the Atomic Simulation Environment (ASE)~\cite{larsen2017atomic,bahn2002object}, which allows precise control over structural parameters and simulates dataset. Following the prior work~\cite{kjaer2023deepstruc}, we consider adataset comprising a model system of monometallic nanoparticles(MMNPs) with seven different structure types. Specifically, structural types includes face-centered cubic (FCC), body-centered cubic (BCC), simple cubic (SC), hexagonal close-packed (HCP), icosahedral, decahedral, and octahedral structures. Atomic structure data are first generated using ASE and DiffPy-CMI, after which the corresponding PDFs are computed using DiffPy-CMI. The parameters used for PDF generation are summarized in Table~\ref{tab:data_parameters}. Here, $Delta2$ denotes the contribution coefficient of the term (1/$r^2$) in peak sharpening.


\begin{table}[ht]
\label{tab:data_parameters}
\caption{Parameters for the generation of simulated PDF data.}
\begin{ruledtabular}
\label{tab:data_parameters}
\begin{tabular}{ccccccccc}
Parameter & $r_{\text{min}}$ (\AA) & $r_{\text{max}}$ (\AA) & $r_{\text{step}}$ (\AA) & $Q_{\text{min}}$ (\AA$^{-1}$) & $Q_{\text{max}}$ (\AA$^{-1}$) & $Q_{\text{damp}}$ (\AA$^{-1}$) & $B_{\text{iso}}$ (\AA$^{-2}$) & Delta2 (\AA$^{-2}$) \\ 
Value & 0.0 & 30.0 & 0.01 & $0.7$ & $25.0$ & $ 0 \sim 0.1$ & 0.3 & 0.0 \\ 
\end{tabular}
\end{ruledtabular}
\end{table}

\begin{figure}
\centering 
\includegraphics[width=8.5cm]{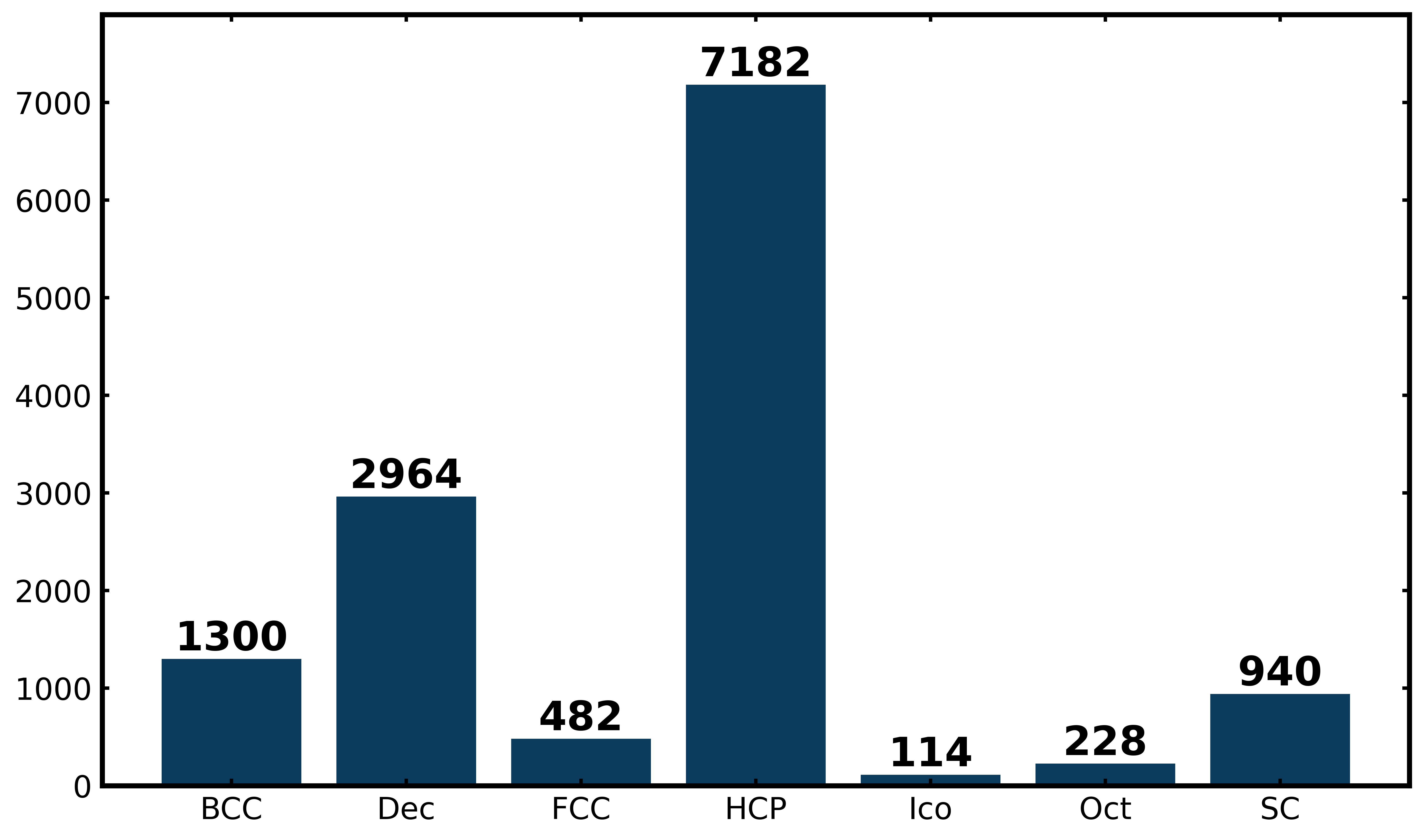}
\caption{The simulated dataset generated seven structure types of data structures, and the number of different types is shown in the figure.}
\label{fig:Struc} 
\end{figure}

All structures considered in this study contain between 5 and 256 atoms. In total, 13,210 unique structures are included in the dataset, and the number of each type in the generated dataset is shown in Fig.~\ref{fig:Struc}. These data are used to construct the training and validation datasets for evaluating the feasibility of the proposed CbLDM framework. A fixed random seed ($42$) is used to split the dataset into 95\% training dataset and 5\% validation dataset, ensuring reproducibility of the results. The training dataset is used to fit model parameters, while the validation dataset provides an independent check on model behavior under the same data distribution. In addition, to empirically explore the feasible interval between $T_2$ and $T_1$ discussed in Section~\ref{sampling step}, a small auxiliary dataset consisting of 36 BCC and 19 SC structures is generated.
In addition to the simulated data used for training and validation as described above, an experimental dataset is introduced exclusively as a test dataset. This test dataset is used solely to assess the final generative capability of the trained model and is never involved in parameter optimization or hyperparameter selection. The experimental PDF data are taken from the study reported in~\cite{kjaer2023deepstruc}. In total, five experimental PDF datasets are considered, for which no corresponding atomic structure information is available.

To quantitatively assess the generation quality of the model, the $R_{wp}$ is adopted as an evaluation metric. $R_{wp}$ is defined by this equation: $ R_{wp} = \{ \frac{\sum w_i [ Y_i(obs) - Y_i(calc) ] ^2}{\sum w_i [ Y_i(obs) ] ^2 } \} ^ {\frac{1}{2}} $, where $w_i$ represents the weight corresponding to position i, $Y_i(obs)$ represents simulation generated PDF for training, $Y_i(calc)$ represents model predicted PDF. The $R_{wp}$ value reflects the relative residual between the two PDFs and is widely used as a reasonable indicator of evaluation. In this study, uniform weights ($w_i = 1$) are adopted to simplify the evaluation, meaning that $R_{wp} = \{ \frac{\sum  [ Y_i(obs) - Y_i(calc) ] ^2}{\sum [ Y_i(obs) ] ^2 } \} ^ {\frac{1}{2}}$.

\subsection{Results on simulated data}
\begin{figure}[htbp]
\centering 
\includegraphics[width=8.5cm]{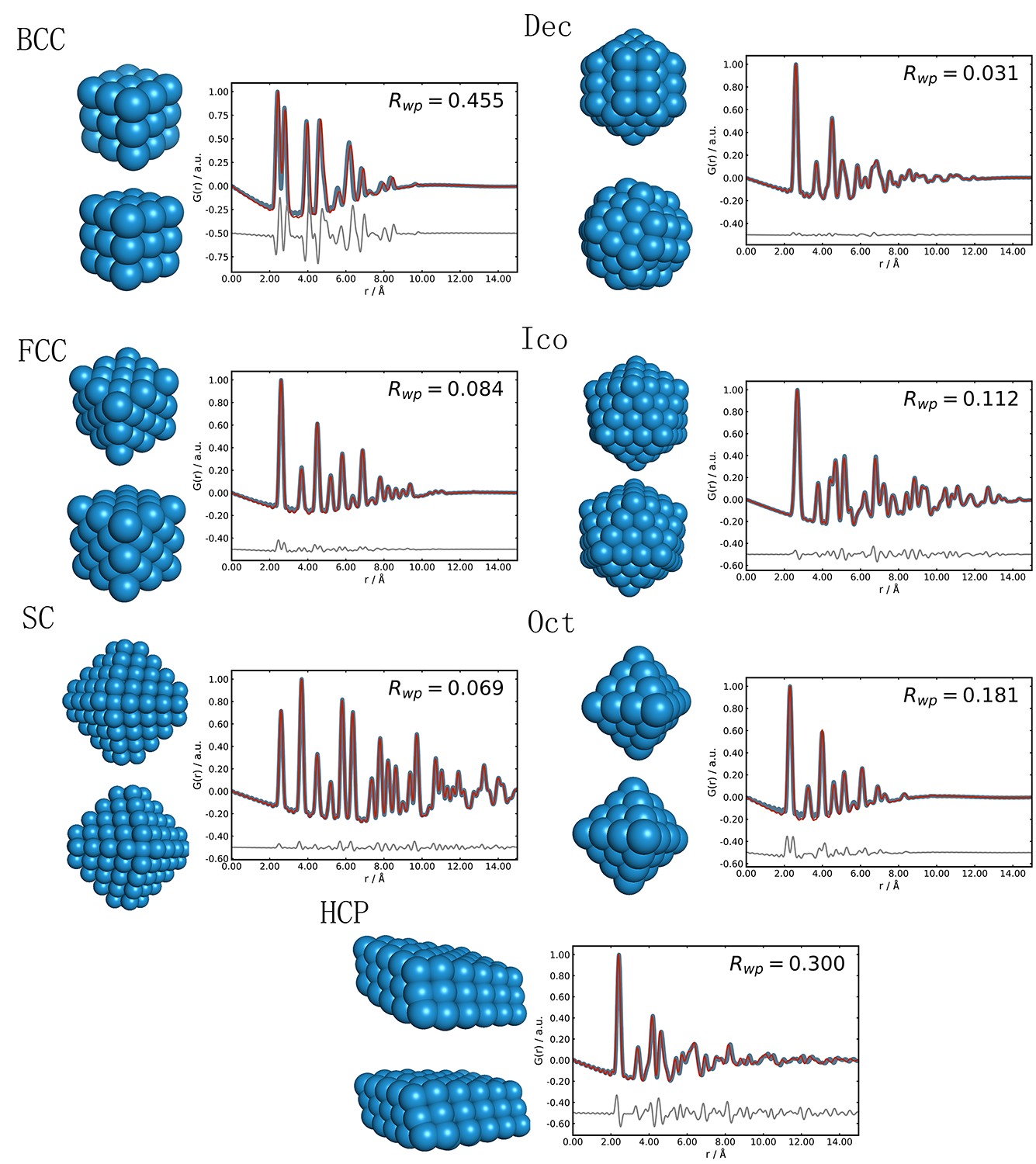}
\caption{Representative results of CbLDM on simulated data from seven structural types. Each unit consists of three components: the upper-left image shows the reference atomic structure, the lower-left image shows a structure generated by CbLDM, and the right image compares the corresponding PDFs, together with the residual curve and the associated $R_{wp}$ value. The blue curve denotes the reference PDF, while the red curve denotes the PDF calculated from the generated structure. For clarity, the residual curve is vertically shifted downward.} 
\label{fig:seven structures} 
\end{figure}

As illustrated in Fig.~\ref{fig:seven structures}, representative examples from each of the seven structural types are selected to qualitatively demonstrate the behavior of CbLDM on simulated training dataset. As shown in Fig.~\ref{fig:seven structures}, the generated atomic structures exhibit similiarity with the corresponding reference structures, and the resulting PDFs yield relatively small $R_{wp}$ values. For the BCC case, the overall structural nature is correctly captured, although noticeable fluctuations are observed in the corresponding PDF residual curves. This behavior is likely attributable to the sensitivity of the PDF fitting procedure to specific parameter choices, which can introduce small horizontal shifts between the calculated and reference PDFs.

Subsequently, for the seven structural types shown in Fig.~\ref{fig:seven structures}, we conduct a comparative study involving several representative baseline approaches, including MLP, CNN~\cite{lecun1998gradient}, ResNet~\cite{he2016deep}, Transformer~\cite{vaswani2017attention}, DeepStruc (CVAE)~\cite{kjaer2023deepstruc}, MLstructureMining (XGBoost)~\cite{kjaer2024mlstructuremining}, together with the proposed CbLDM, as summarized in Tab.~\ref{tab:Rwp}. Among these, DeepStruc utilised a pre-trained model, while the remaining models are trained solely on our simulated dataset. The number of parameters in each model was broadly comparable to that of CbLDM. That is to say, for DeepStruc, this data served as validation data. Nevertheless, Tab.~\ref{tab:Rwp} also demonstrates that CbLDM exhibits superior adaptability to this problem compared to the other models. For each structure type, 100 samples are randomly drawn from the training dataset to provide a representative qualitative and quantitative comparison. In addition, 1000 samples are randomly selected from the full training dataset to obtain an aggregated reference across structural types, which is reported in the ``total'' row of Tab.~\ref{tab:Rwp}. The results indicate that CbLDM achieves reconstructions of training dataset, demonstrating feasibility and stability comparable to existing approaches.

\begin{table}[ht]
\label{tab:Rwp}
\caption{$R_{wp}$ values of different models on seven different structure types.}
\renewcommand{\arraystretch}{1.2}
\begin{ruledtabular}
\label{tab:Rwp}
\begin{tabular}{cccccccc}

& MLP & CNN & ResNet & Transformer & DeepStruc & MLstructureMining & CbLDM \\ 
bcc & 1.512 & 1.373 & 1.358 & 1.584 & 0.793 & 1.196 & \textbf{0.410} \\ 
dec & 1.248 & 1.197 & 1.214 & 1.249 & 0.692 & 1.116 & \textbf{0.026} \\ 
fcc & 1.318 & 1.240 & 1.286 & 1.350 & 1.682 & 1.221 & \textbf{0.068} \\
hcp & 1.234 & 1.210 & 1.152 & 1.244 & 1.347 & 1.145 & \textbf{0.263} \\ 
ico & 1.349 & 1.298 & 1.273 & 1.366 & 1.605 & 0.459 & \textbf{0.095} \\
oct & 1.344 & 1.277 & 1.267 & 1.359 & 1.590 & 1.156 & \textbf{0.151} \\
sc  & 1.431 & 1.328 & 1.346 & 1.478 & 1.107 & 0.808 & \textbf{0.059} \\ 
total & 1.278 & 1.228 & 1.191 & 1.299 & 1.266 & 1.103 & \textbf{0.380} \\ 
\end{tabular}
\end{ruledtabular}
\end{table}

From Tab.\ref{tab:Rwp}, it can be observed that CbLDM consistently produces recovery of nanostructure that are across all considered structure types. Among the cases, the decahedral structure provides a representative example, for which CbLDM yields a low Rwp value of 0.026, indicating close similiarity between the generated and reference PDFs.

\begin{figure}[htbp]
\centering 
\includegraphics[width=8.5cm]{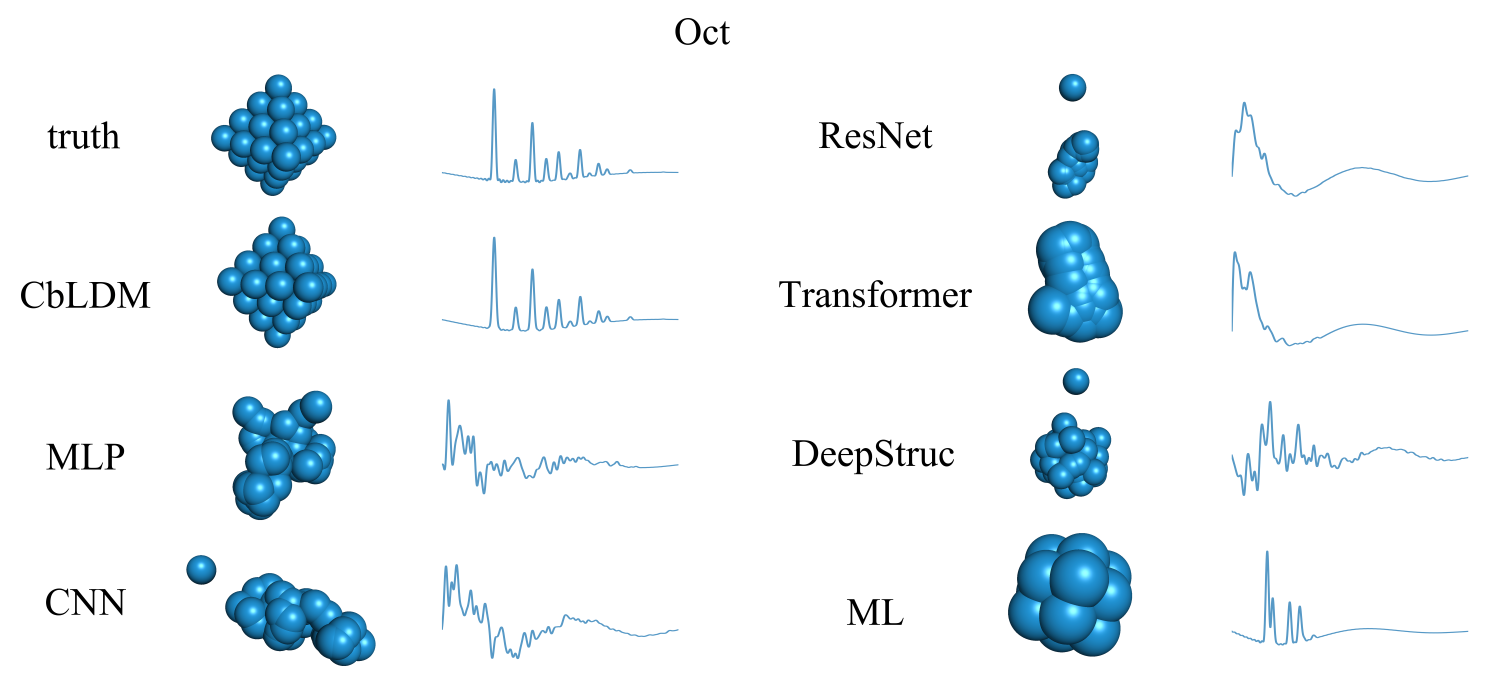}
\caption{The prediction results of different models on the validation dataset for the Oct structure are illustrated in this figure. From top to bottom and from left to right, the figure presents the ground-truth Oct structure, followed by the predictions generated by CbLDM, MLP, CNN, ResNet, Transformer, DeepStruc (CVAE), and MLstructureMining (XGBoost). The first and fourth columns indicate the corresponding model names, the second and fifth columns display the reconstructed atomic structures, and the third and sixth columns show the associated PDFs.} 
\label{fig:Oct_compare} 
\end{figure}

To further illustrate the generalization of the proposed framework, representative samples from the validation dataset are selected for qualitative visualization across the seven structure types. As shown in Figure \ref{fig:Oct_compare}, the first row of the first column displays the reference atomic structure and its corresponding PDF, while the remaining rows present the structures and PDFs generated by different models. One Oct structure is selected as a representative case for visualization on the validation dataset, while the corresponding results for the remaining structure types are provided in the supplementary material.

Additionally, we compared CbLDM with DeepStruc on the validation dataset, with results shown in Tab.~\ref{tab:Rwp on validation}. As can be seen from the Tab.~\ref{tab:Rwp on validation}, CbLDM maintains equivalent predictive capability in terms of generalisation performance for MMNPs.

\begin{table}[ht]
\label{tab:Rwp on validation}
\caption{$R_{wp}$ values of DeepStruc and CbLDM on validation dataset.}
\renewcommand{\arraystretch}{1.2}
\begin{ruledtabular}
\label{tab:Rwp on validation}
  \begin{tabular}{ccccccccc}

    Atoms & & bcc & dec & fcc & hcp & ico & oct & sc \\
    \multirow{2}{*}{5--80} & CbLDM & 0.389 & 0.611 & 0.738 & 0.673 & 1.096& 0.865& 0.614\\
    & DeepStruc & 1.606& 1.810& 1.684& 1.753& 1.472& 1.831& 1.904\\
    \multirow{2}{*}{80--170} & CbLDM & 0.302& 0.217& 0.237& 0.546& --& 0.368& 0.213\\
    & DeepStruc & 1.433& 1.692& 1.663& 1.822& --& 1.464& 1.471\\
    \multirow{2}{*}{170--256} & CbLDM & 0.659& 0.191& 0.207& 0.544& --& 0.506& 0.316\\
    & DeepStruc & 1.459& 1.754& 1.541& 1.783& --& 1.762& 1.522\\

  \end{tabular}
\end{ruledtabular}
\end{table}

\begin{figure}[htbp]
\centering 
\includegraphics[width=8.5cm]{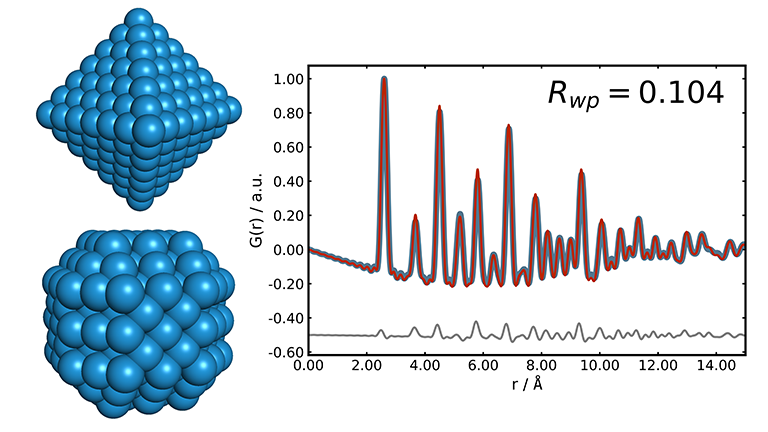}
\caption{The phenomenon of two seemingly different structures with the similar PDFs. The top left and bottom left are the two atomic structures, and the right is the comparison of PDFs, the residual and $R_{wp}$.} 
\label{fig:similar} 
\end{figure}

During inference, we observe that the model can generate seemingly distinct atomic structures whose corresponding PDFs are highly similar, as illustrated in Fig.\ref{fig:similar}. Both structures correspond to Oct structure generated by CbLDM. This behavior can be explained by the fact that the two structures share similar internal stacking arrangements while differing in their cross section geometries. This observation suggests that CbLDM captures the relationship between PDF information and local structural features in a consistent manner. More fundamentally, this phenomenon arises from intrinsic limitations of PDF data, rather than deficiencies of the model itself. Nevertheless, the CbLDM is able to produce multiple plausible structural reconstructions from a single PDF, providing a broader set of candidate solutions.

\subsection{Results on experimental data}
\begin{figure}[htbp]
\centering 
\includegraphics[width=8.5cm]{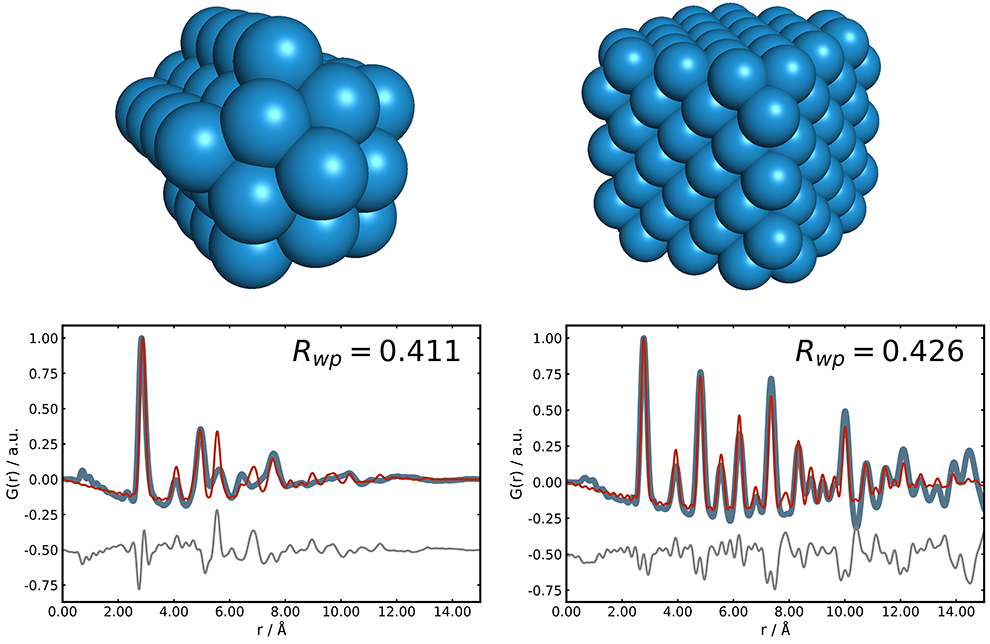}
\caption{Prediction results of CbLDM on the test dataset. Two representative samples are presented, where the generated atomic structures are shown, alongside a comparison between the PDFs calculated from the generated structures and the input PDFs.} 
\label{fig:exPDF} 
\end{figure}

As shown in Fig.\ref{fig:exPDF}, we present two representative experimental cases, namely $Au_{144}(p-MBA)_{60}$ and $Pt$ nanoparticle, to qualitatively examine the applicability of CbLDM to real PDF data. The true structure of $Au_{144}(p-MBA)_{60}$ is the Dec structure and the true structure of $Pt$ nanoparticle is the FCC structure. For both cases, the structures generated by CbLDM yield PDFs that are qualitatively consistent with the corresponding experimental inputs. In both cases, the structures generated by CbLDM produce PDFs that are in agreement with the corresponding experimental measurements. In comparison with DeepStruc, CbLDM is able to produce structurally plausible candidates using fewer generation times in these examples, suggesting a potential efficiency advantage for conditional sampling. The figure shows the atomic structures generated by CbLDM for $Au_{144}(p-MBA)_{60}$ and $Pt$ nanoparticle, together with a comparison between the experimental PDFs and those calculated from the generated structures. The left side corresponds to the $Au_{144}(p-MBA)_{60}$, while the right side shows results for the $Pt$ nanoparticle.

\section{Conclusion and Discussion}

This work explores the feasibility of addressing the nanostructure inverse problem from PDF data within a probabilistic generative model framework. Therefore, we propose a CbLDM as a proof-of-concept approach for conditional structure generation. CbLDM departs from the conventional VAE formulation by replacing the standard normal prior with a conditional prior, aiming to promote continuity and consistency in the latent space with respect to condition information. In addition, CbLDM can reduce computational costs and save some time by utilizing conditional prior distributions during the sampling process. Subsequently, DM is employed to denoise the noisy coarse samples, obtaining refined samples within the latent space based on conditions. Furthermore, employing the Laplacian matrix in place of the direct distance matrix as the model output representation attenuate the influence of long-range uncertainty in the recovery of nanostructures. The experimental results serve to demonstrate the feasibility of the proposed framework, showing that CbLDM is capable of generating physically meaningful nanostructure representations consistent with PDF data.

This study focuses on the investigation of MMNPs, researching on nanostructure recovery based on CbLDM. It lays a solid foundation for subsequent efforts to solve complex inverse problems about such as the model system of polymetallic nanoparticles and nested nanostructures. As a proof-of-concept study, this work naturally opens several directions for future investigation. For example, the model focuses solely on accelerating conditional generation during sampling. Exploring strategies for incorporating conditional information randomly for generation remains an open question. Future work may investigate principled ways of embedding unconditional information to influence the sampling process more effectively. Moreover, the current study primarily relies on simulation datasets, and extending the framework to larger collections of experimentally measured PDFs represents an important next step. Such extensions would enable a more comprehensive assessment of the model’s applicability across diverse experimental scenarios. Finally, systematic ablation studies could further clarify the roles of individual components within CbLDM and inform subsequent methodological refinements.

\begin{acknowledgments}
This work was supported by the National Key R\&D Program of China No.2022YFA1003800, the Natural Science Foundation of Tianjin No. 25JCJQJC00300, and the Fundamental Research Funds for the Central Universities No.63253105 for RG. Work in the Billinge group (SJLB and LL) was carried out with support from  U.S. Department of Energy, Office of Science, Office of Basic Energy Sciences (DOE-BES) grant contract No. DE-SC0024141.
\end{acknowledgments}
\section*{AUTHOR DECLARATIONS}
\subsection*{Conflict of Interest}
The authorw have no conflicts to disclose.

\subsection*{Author Contributions}
Jiarui Cao: Conceptualization (equal); Methodology (equal); Software (equal); Validation (equal); Formal analysis (equal); Writing-original draft (equal); Writing-review \& editing (equal). Zhiyang Zhang: Conceptualization (equal); Methodology (equal); Software (equal); Validation (equal); Formal analysis (equal); Writing-original draft (equal); Writing-review \& editing (equal). Heming Wang: Conceptualization (equal); Methodology (equal); Software (equal); Validation (equal); Formal analysis (equal); Writing-original draft (equal); Writing-review \& editing (equal). Jun Xu: Data curation (supporting); Writing-review (supporting). Ling Lan: Investigation (supporting); Writing-review \& editing (supporting). Simon J. L. Billinge: Conceptualization (supporting); Writing-review \& editing (supporting). Ran Gu: Conceptualization (lead); Methodology (lead); Supervision (lead); Funding acquisition (lead); Writing-original draft (lead); Writing-review \& editing (lead).
\section*{DATA AVAILABILITY}
The code and compressed data needed to reproduce the findings of this study are openly available at \url{https://huggingface.co/wszzyang/CbLDM} and \url{https://github.com/hegwj/CbLDM}.
\appendix
\section{The Pair Distribution Function (PDF)}
\begin{figure}
    \centering
    \includegraphics[width=8cm]{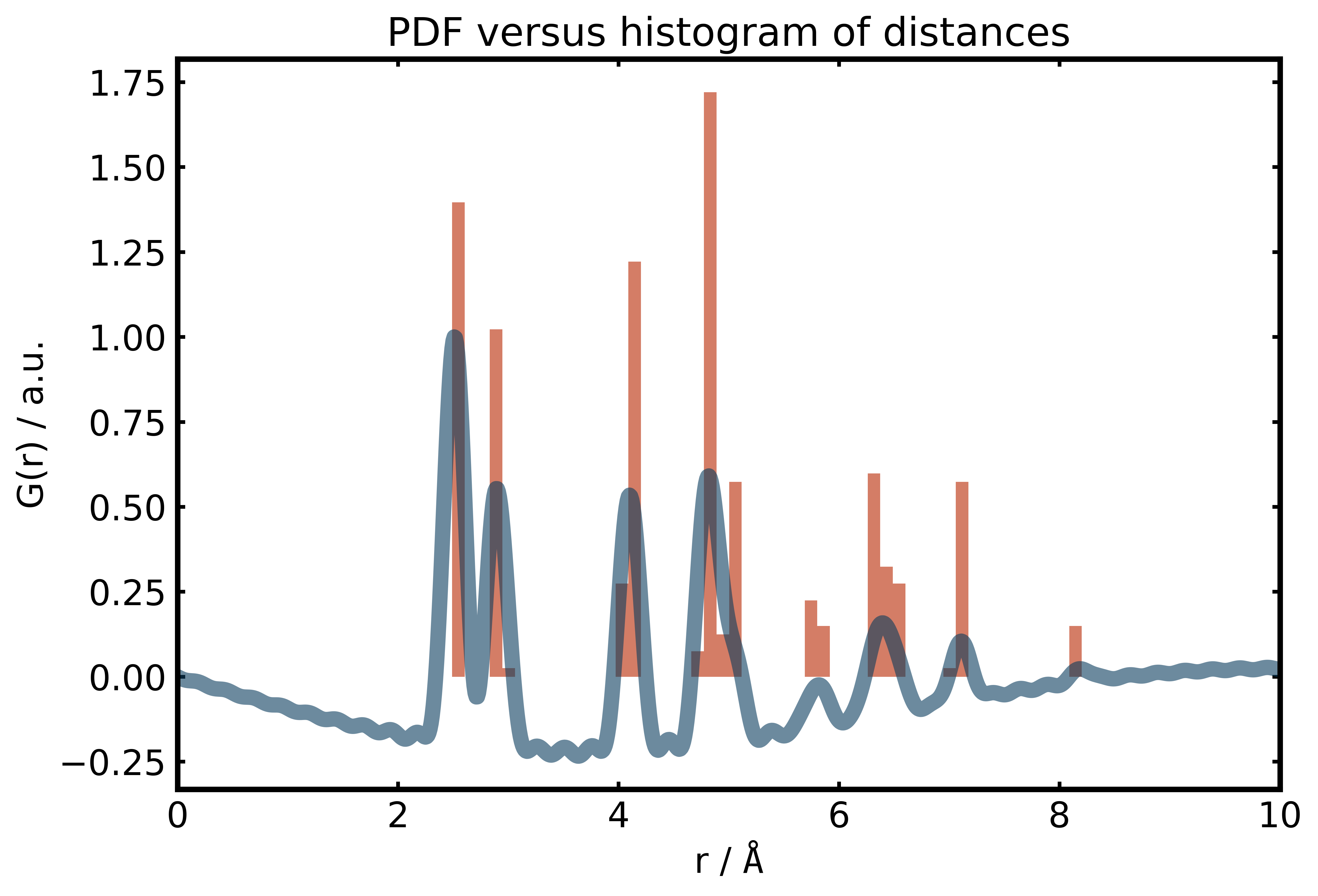}
    \caption{
    The pair distribution function (PDF) $G(r)$ versus interatomic distances $r($\AA$)$, alongside a histogram of atomic distances. The blue curve represents the PDF, while the orange curve shows the histogram of interatomic distances, calculated every 0.2~\AA. This simulated PDF is derived from a specified structure body centered cubic (BCC).} 
    \label{fig:pdf and hist} 
\end{figure}

The pair distribution function (PDF), derived from Fourier transform of normalized and corrected experimental X-ray total scattering data~\cite{proffen2003structural,neder2005structure}, is a powerful and widely used tool for nanoscale structural analysis~\cite{egamiBraggPeaksStructural2012, billinge2004beyond}. Theoretically, it provides a one-dimensional measure of the probability of finding pairs of atoms separated by a distance $r($\AA$)$ in the powder sample~\cite{billingeLocalStructureTotal2008}. Peaks in the PDF correspond to distances with high probability, and as illustrated in Fig.~\ref{fig:pdf and hist}, these peaks closely align with a histogram of all interatomic distances in the sample. Consequently, it is possible to extract a list of distances directly from the PDF. This approach has been successfully implemented by fitting the PDF with a model based on the Debye scattering equation~\cite{granlund2015algorithm,gu2019algorithm}.
\section{The Distance Geometry Problem}
Using the distance information in the PDF to find the chemical structure can be regarded as an unassigned distance geometry problem (uDGP) ~\cite{billinge2016assigned}. To be more specific, it involves finding a set of vertices given the distance list, such that the distances between the vertices correspond one-to-one with the target distance list \cite{duxbury2016unassigned}. Mathematically, uDGP is equivalent to solving a mapping problem between a set (list of interatomic distances) and a graph structure (atomic structure), ultimately mapping the graph structure to three-dimensional coordinates. 

If the graph structure of uDGP is given, meaning that the vertex information of each distance is known, it degenerates into the assigned distance geometry problem (aDGP), which is widely used in protein folding and sensor localization problems~\cite{more1999distance,so2007theory}. Although both uDGP and aDGP are NP-hard~\cite{kjaer2023deepstruc}, intuitively uDGP is considered harder because it involves an additional process of assigning edges compared to aDGP, and finding the correct assignment is challenging. Traditional methods like TRIBOND and LIGA incrementally add new atoms on a small structure, incorporating the methods of aDGP in the process of adding new atoms, which involves trial and error leading to frequent backtracking and significant computational costs. The DeepStruc model also considers this uDGP problem, but instead of solving the uDGP directly, it uses a data-driven method to learn the adjacency matrix $A \in R^{N\times N}$, where the entries of the adjacency matrix are the Euclidean distance between each pair of the $N$ atoms. Predicting the adjacency matrix is machine-learning friendly as it can overcome the non-uniqueness of solutions caused by isometric transformations like rotations and translations. The authors then transformed the distance matrix into three-dimensional coordinates by solving the aDGP subproblem. 

In this study, our CbLDM approach follows a procedure similar to DeepStruc in that it first transforms the uDGP into an aDGP, after which an aDGP method is applied to recover the three-dimensional coordinates, thereby overcoming the non-uniqueness of solutions caused by isometric transformations such as rotations and translations. Unlike DeepStruc, however, our method replaces the conventional distance matrix in the structural representation graph G = (X, A) with a Laplacian matrix, which mitigates error propagation from noisy long-range measurements by assigning lower weights to large interatomic distances.

\section{Laplacian Matrix}
For an undirected graph $G(V,E)$ with $n$ vertices, we consider a fully connected graph here because the PDF covers distance information for all pairs of atoms, although the information for large distances is relatively inaccurate. The Laplacian matrix~\cite{merris1994laplacian,lee2008clustering} is defined by the weighted degree matrix $D$ and the adjacency matrix $W$ as $L=D-W$ with
\begin{equation}
    W_{ij} = \texttt{exp}(-\frac{\|v_i - v_j\|^2_2}{2\sigma^2}) \text{ and } 
    D_{ii} = \sum_{j=1}^n W_{ij},
    \label{matrixwdefine}
\end{equation}
where $v_i$ and $v_j$ are the vertices of the nanostructure and $\sigma$ is a parameter. From the definition~\eqref{matrixwdefine}, we know that the larger the distance between $v_i$ and $v_j$, the smaller $L_{ij}$ is. Therefore, compared to the distance matrix, in Laplacian matrix, small distances are less affected by errors, making it more suitable for PDF data.

In addition, another property of the Laplacian matrix facilitates the subsequent solution of aDGP. For the three-dimensional embedding problem of graph $G$, where nearby vertices should remain as close as possible after mapping, this can be formulated as the following optimization problem
\begin{equation}
    \min \mathop{\rm{\sum}}_{
        i,j} W_{ij}\Vert z_i - z_j \Vert^2_2.
    \label{laplace1}
\end{equation}
The solution to problem \eqref{laplace1} is equivalent to solving the generalized eigenvalue problem
$$Ly= \lambda Dy,$$
where the eigenvectors corresponding to the three smallest non-zero generalized eigenvalues form the solution. Although this solution is not obtained in terms of mean squared error (MSE), meaning that the Laplacian matrix generated by these three-dimensional coordinates may not be the closest to the target, it still retains certain structural information and serves as a useful initial guess for the aDGP.
\section{Latent Diffusion Model}
LDM(Latent Diffusion Model)~\cite{rombach2022high} is a probabilistic generative model, with its diffusion and denoising processes in the latent space. The probabilistic generation model also coincides with the fact that the atomic structure corresponding to the PDF is probabilistic. The latent space can be used to sample the corresponding Laplacian matrix based on the information provided by PDF. We review the principles of LDM in this section, which we present in two parts: the VAE and the DDM(Denoising Diffusion Model).
\subsection{Variational Autoencoders}
VAE(Variational Autoencoder) is an extension of Autoencoder. It introduces the concept of probability to the traditional Autoencoder, which enables the model to learn the latent distribution of the data and thus has the ability to generate new data~\cite{kingma2013auto,girin2020dynamical}. VAE contains two parts, encoder and decoder. VAE implements the compression process of the data through an Encoder to obtain the posterior distribution $q_\theta(z|x)$, and the decompression process of the data through a Decoder to obtain the conditional distribution $p_\theta(x|z)$, which represents the probability of generating the observation given the latent variable. $x$ represents ground truth data, $z$ represents latent data. VAE assumes the existence of a latent variable z that controls the data generation process whose joint probability distribution can be decomposed as:
\begin{equation*}
    p_\theta(x,z) = p_\theta(x|z)p(z)
    \label{vae xz distribution}
\end{equation*}
where $p(z)$ presents prior distribution of $z$. Here, we usually assume that $p(z)$ is $N(0,I)$. The loss function of VAE consists of two parts: the reconstruction loss and the KL-divergence, as shown in Eq.~\eqref{vae_loss}:
\begin{align}
    L(\theta) &= -E_{q_\theta(z|x)}[\log p_\theta(x|z)] \notag\\&+ D_{KL}(q_\theta(z|x)||p_\theta(z))
    \label{vae_loss}
\end{align}
where $L$ represents the loss function, $\theta$ represents the parameters learnt by the model, and $ D_{KL}(q_\theta(z|x)||p(z)) = \int q_\theta(z|x)\log \frac{q_\theta(z|x)}{p(z)}\,dz$.
\subsection{Denoising Diffusion Models}
DDM(Denoising Diffusion Model) is a generation model based on Markov chain. The working principle of DDM includes two main processes: diffusion process and denoising process.~\cite{ho2020denoising} Diffusion process gradually adds Gaussian noise to real data $x_0$ to obtain a final pure Gaussian noise $x_T$. In this process, a series of noisy data with different degrees of noise $x_1,x_2\cdots x_T$ can be generated. These noisy data and the added Gaussian noise provide a large number of data samples for model training. We assume that the moments are only related to moments throughout the process, and thus the entire diffusion process constitutes a Markov chain. And since its state transfer follows a Gaussian distribution, the Markov chain is called a Gaussian Markov chain. The method of adding noise during diffusion is as Eq.~\eqref{noisy method1}.
\begin{equation}
    x_t = \sqrt{1-\beta_t}x_{t-1}+\beta_t \epsilon \quad \epsilon \sim N(0,I)
    \label{noisy method1}
\end{equation}
where $\beta_t$ is a small positive value, representing the value of variance of noise added in time step $t$. After a series of derivation, we can get the Eq.~\eqref{noisy method2}:
\begin{align}
    x_t &= \sqrt{(1-\beta_1)(1-\beta_2)\cdots(1-\beta_t)}x_{0}\notag \\
    &+\sqrt{1-(1-\beta_1)(1-\beta_2)\cdots(1-\beta_t)} \epsilon
    \label{noisy method2}
\end{align}
let $\alpha = 1-\beta$,$\overline{\alpha_t} = (1-\beta_1)(1-\beta_2)\cdots(1-\beta_t)$ and have:
\begin{equation}
    x_t = \sqrt{\overline{\alpha_t}}x_{0}+\sqrt{1-\overline{\alpha_t}} \epsilon \quad \epsilon \sim N(0,I)
    \label{noisy method3}
\end{equation}
Although theoretically, the process of adding noise from $x_0$ to $x_t$ requires $t$ time steps of noise addition. In fact, it can be obtained in one step from $x_0$ to $x_t$ according to the Eq.\eqref{noisy method3}. The denoising process of DDM can be understood as the process of generating $x_{t-1}$ given $x_t$, iteratively generating the final $x_0$ through Markov chain, and it is also the process that the model needs to learn. The loss function of DDM is represented as Eq.\eqref{ddm loss}.
\begin{align}
L &=  E_{q(x_1|x_0)}[-\log_{p_\theta}(x_0|x_1)]\notag\\
&+\sum^T_{t=2}D_{KL}(q(x_{t-1}|x_t,x_0)||p_\theta(x_{t-1}|x_t)) \notag\\
    &+D_{KL}(q(x_T|x_0)||p_\theta(x_T)) 
\label{ddm loss}
\end{align}
where $p_\theta(\cdot)$ represents the distribution of denoising process that the neural network needs to learn. The third term is a fixed constant that is independent of $\theta$, so we don't need to pay attention to it. The first two terms can be further calculated and simplified~\cite{sohl2015deep}, resulting in a simplified objective function:
\begin{equation*}
    L_{simple} = E_{t,x_0,\epsilon}[||\epsilon - \epsilon_\theta(\sqrt{\overline{\alpha_t}}x_0+\sqrt{1-\overline{\alpha_t}}\epsilon,t)||^2]
\end{equation*}
where the $\epsilon$ represents the real noise added to the data during the diffusion process, $\epsilon_\theta(\sqrt{\overline{\alpha_t}}x_0+\sqrt{1-\overline{\alpha_t}}\epsilon,t)$ represents the noise predicted by the reverse denoising network, and $t$ is sampled from $\{1,...,T\}$.

In summary, we have introduced VAE and DDM, which together comprise LDM. LDM performs diffusion and denoising on the latent space encoded by VAE, which greatly reduces the computational and memory occupation, and improves the generation efficiency compared to processing directly in the original high-dimensional data space.
\nocite{*}
\bibliography{aipsamp}

\end{document}